\documentclass{article} 
\usepackage{iclr2020_conference,times}

\usepackage{amsmath,amsfonts,bm}

\def\eqref#1{equation~\ref{#1}}

\def\1{\bm{1}}

\DeclareMathAlphabet{\mathsfit}{\encodingdefault}{\sfdefault}{m}{sl}
\SetMathAlphabet{\mathsfit}{bold}{\encodingdefault}{\sfdefault}{bx}{n}

\newcommand{\indep}{\mbox{${}\perp\mkern-11mu\perp{}$}}
\newcommand{\notindep}{\mbox{${}\not\!\perp\mkern-11mu\perp{}$}}

\newcommand{\KL}{D_{\mathrm{KL}}}

\usepackage{hyperref}
\usepackage{url}
\usepackage{amsmath,amssymb,amsfonts,fixmath,siunitx,bm,graphicx,color,caption,tikz,booktabs,adjustbox,listings}
\usepackage[utf8]{inputenc} 
\usepackage[T1]{fontenc}    
\usepackage{hyperref}       
\usepackage{url}            
\usepackage{booktabs}       
\usepackage{nicefrac}       
\usepackage{microtype}      
\usepackage{bm}

\def\fdA{./figures/}

\newcommand{\ig}[3]{\includegraphics[width=#1\textwidth,type=#2,ext=.#2,read=.#2]{#3}}

\def\trVAE{
\begin{figure}[t!]
\centering
\ig{0.9}{pdf}{\fdA Figure1}
\caption{The transformer VAE (trVAE) is an MMD-regularized CVAE. It receives randomized batches of data ($x$) and condition ($s$) as input during training, stratified for approximately equal proportions of $s$. In contrast to a standard CVAE, we regularize the effect of $s$ on the representation obtained after the first-layer $g_1(\hat z, s)$ of the decoder $g$. During prediction time, we transform batches of the source condition $x_{s=0}$ to the target condition $x_{s=1}$ by encoding $\hat z_0 = f(x_0, s=0)$ and decoding $g(\hat{z}_0, s=1)$.}
\label{fig:trVAE}
\end{figure}}

\def\latent{
\begin{figure}[t!]
\centering
\ig{0.7}{pdf}{\fdA Figure6}
\caption{Comparison of representations for MMD-layer in trVAE and the corresponding layer in the vanilla CVAE using UMAP \citep{2018arXivUMAP}. The MMD regularization incentivizes the model to learn condition-invariant features resulting in a more compact representation. The figure shows the qualitative effect for the ``PBMC data'' introduced in experiments section \ref{sc:Kang}. Both representations show the same number of samples.}
\label{fig:latent}
\end{figure}}

\def\Haber{
\begin{figure}[t!]
\centering
\ig{1.0}{pdf}{\fdA Figure4}
\caption{\textbf{(a)} UMAP visualization  of conditions and cell type for gut cells. \textbf{(b-c)} Mean and variance expression of 1,000 genes comparing trVAE-predicted and real infected Tuft cells together with the top 10 differentiall-expressed genes highlighted in red ($R^2$ denotes Pearson correlation between ground truth and predicted values). \textbf{(d)} Distribution of \textit{Defa24}: the top response gene to H.poly.Day10 infection between control, predicted and real stimulated cells for different models. Vertical axis: expression distribution for \textit{Defa24}. Horizontal axis: control, real and predicted distribution by different models. \textbf{(e)} Comparison of Pearson's $R^2$ values for mean and variance gene expression between real and predicted cells for different models. Center values show the mean of $R^2$ values estimated using $n = 100$ random subsamples for the prediction of each model and error bars depict standard deviation. \textbf{(f)} Comparison of $R^2$ values for mean gene expression between real and predicted cells by trVAE for the eight different cell types and three conditions. Center values show the mean of $R^2$ values estimated using $n = 100$ random subsamples for each cell type and error bars depict standard deviation.}
\label{fig:Haber}
\end{figure}}

\def\Kang{
\begin{figure}[t!]
\centering
\ig{1.0}{pdf}{\fdA Figure5}
\caption{\textbf{(a)} UMAP visualization of peripheral blood mononuclear cells (PBMCs). \textbf{(b-c)} Mean and variance per 2,000 dimensions between trVAE-predicted and real natural killer cells (NK) together with the top 10 differentially-expressed genes highlighted in red. \textbf{(d)} Distribution of \textit{ISG15}: the most strongly changing gene after IFN-$\beta$ perturbation between control, real and predicted stimulated cells for different models. Vertical axis: expression distribution for \textit{ISG15}. Horizontal axis: control, real and predicted distribution by different models. \textbf{(e)} Comparison of $R^2$ values for mean and variance gene expression between real and predicted cells for different models. Center values show the mean of $R^2$ values estimated using $n = 100$ random subsamples for the prediction of each model and error bars depict standard deviation.}
\label{fig:Kang}
\end{figure}}

\def\MNIST{
\begin{figure}[t!]
\centering
\ig{0.6}{pdf}{\fdA Figure2}
\caption{Out-of-sample style transfer for Morpho-MNIST dataset containing normal, thin and thick digits. trVAE successfully transforms normal digits to thin \textbf{(a)} and thick (\textbf{(b)} for digits not seen during training (out-of-sample).}
\label{fig:mnist}
\end{figure}}

\def\CelebA{
\begin{figure}[t!]
\centering
\ig{0.7}{pdf}{\fdA Figure3}
\caption{CelebA dataset with images in two conditions: celebrities without a smile and with a smile on their face. trVAE successfully adds a smile on faces of women without a smile despite these samples completely lacking from the training data (out-of-sample). The training data only comprises non-smiling women and smiling and non-smiling men.}
\label{fig:celeba}
\end{figure}}

\def\tabletrVAE{
\begin{table}[h!]
\caption{trVAE detailed architecture. We used the same architecture for all the examples in the paper. The input\_dim parameter for each dataset is: IFN‐$\beta$ (2,000), H.poly (1,000).}
\label{tab:trVAE}
\begin{center}
\begin{tabular}{lllllll}
\multicolumn{1}{c}{\bf Name}  &\multicolumn{1}{c}{\bf Operation}&\multicolumn{1}{c}{\bf NoF/Kernel Dim.}&\multicolumn{1}{c}{\bf Dropout}&\multicolumn{1}{c}{\bf BN}&\multicolumn{1}{c}{\bf Activation}&\multicolumn{1}{c}{\bf Input}\\ \hline \\ 
input & - & input\_dim & $\times$ & $\times$ & - & - \\
conditions & - & n\_conditions & $\times$ & $\times$ & - & - \\ 
FC-1 & FC & 800 & 0.2 & $\surd$ & Leaky ReLU & [input, conditions] \\
FC-2 & FC & 800 & 0.2 & $\surd$ & Leaky ReLU & FC-1  \\
FC-3 & FC & 128 & 0.2 & $\surd$ & Leaky ReLU & FC-2  \\
mean & FC & 50 & $\times$ & $\times$ & Linear & FC-3 \\
var & FC & 50 & $\times$ & $\times$ & Linear & FC-3 \\
z-sample & FC & 50 & $\times$ & $\times$ & Linear & [mean, var] \\
MMD & FC & 128 & 0.2 & $\surd$ & Leaky ReLU & [z-sample, conditions] \\
FC-4 & FC & 800 & 0.2 & $\surd$ & Leaky ReLU & MMD \\
FC-5 & FC & 800 & 0.2 & $\surd$ & Leaky ReLU & FC-3 \\
output & FC & input\_dim & $\times$ & $\times$ & ReLU & FC-4 \\ 
\hline 
Optimizer & Adam & & & & &  \\
Learning Rate & 0.001 & & & & & \\
Leaky ReLU slope & 0.2 & & & & & \\
Batch Size & 512 & & & & & \\
\# of Epochs & 5000 & & & & & \\
$\alpha$ & 0.00001 & & & & & \\
$\beta$ & 100 & & & & & \\
$\eta$ & 100 & & & & & \\
\end{tabular}
\end{center}
\end{table}
}

\def\tabletrDCVAEMNIST
{
\begin{table}[h!]
\caption{Convolutional trVAE detailed architecture used for Morpho-MNIST dataset.}
\label{tab:trDCVAE_MNIST}
\begin{center}
\begin{tabular}{lllllll}
\multicolumn{1}{c}{\bf Name}  &\multicolumn{1}{c}{\bf Operation}&\multicolumn{1}{c}{\bf NoF/Kernel Dim.}&\multicolumn{1}{c}{\bf Dropout}&\multicolumn{1}{c}{\bf BN}&\multicolumn{1}{c}{\bf Activation}&\multicolumn{1}{c}{\bf Input}\\ \hline \\ 
input & - & (28, 28, 1) & $\times$ & $\times$ & - & - \\ 
conditions & - & 2 & $\times$ & $\times$ & - & - \\ 
FC-1 & FC & 128 & $\times$ & $\times$ & Leaky ReLU & conditions \\
FC-2 & FC & 784 & 0.2 & $\surd$ & Leaky ReLU & FC-1  \\
FC-2\_resh & Reshape & (28, 28, 1) & $\times$ & $\times$ & $\times$ & FC-2 \\
Conv2D\_1 & Conv2D & (4, 4, 64, 2) & $\times$ & $\times$ & Leaky ReLU & [FC-2\_resh, input] \\
Conv2D\_2 & Conv2D & (4, 4, 64, 64) & $\times$ & $\times$ & Leaky ReLU & Conv2D\_1 \\
FC-3 & FC & 128 & $\times$ & $\surd$ & Leaky ReLU & Flatten(Conv2D\_2) \\
mean & FC & 50 & $\times$ & $\times$ & Linear & FC-3 \\
var & FC & 50 & $\times$ & $\times$ & Linear & FC-3 \\
z & FC & 50 & $\times$ & $\times$ & Linear & [mean, var] \\
FC-4 & FC & 128 & $\times$ & $\times$ & Leaky ReLU & conditions \\
FC-5 & FC & 784 & 0.2 & $\surd$ & Leaky ReLU & FC-4  \\
FC-5\_resh & Reshape & (28, 28, 1) & $\times$ & $\times$ & $\times$ & FC-5 \\
MMD & FC & 128 & $\times$ & $\surd$ & Leaky ReLU & [z, FC-5\_resh] \\
FC-6 & FC & 256 & $\times$ & $\times$ & Leaky ReLU & MMD \\
FC-7\_resh & Reshape & (2, 2, 64) & $\times$ & $\times$ & $\times$ & FC-6 \\
Conv\_transp\_1 & Conv2D Transpose & (4, 4, 128, 64) & $\times$ & $\times$ & Leaky ReLU & FC-7\_resh \\
Conv\_transp\_2 & Conv2D Transpose & (4, 4, 64, 64) & $\times$ & $\times$ & Leaky ReLU & UpSampling2D(Conv\_transp\_1) \\
Conv\_transp\_3 & Conv2D Transpose & (4, 4, 64, 64) & $\times$ & $\times$ & Leaky ReLU & Conv\_transp\_2 \\
Conv\_transp\_4 & Conv2D Transpose & (4, 4, 2, 64) & $\times$ & $\times$ & Leaky ReLU & UpSampling2D(Conv\_transp\_3) \\
output & Conv2D Transpose & (4, 4, 1, 2) & $\times$ & $\times$ & ReLU & UpSampling2D(Conv\_transp\_4) \\

\hline
Optimizer & Adam & & & & &  \\
Learning Rate & 0.001 & & & & & \\
Leaky ReLU slope & 0.2 & & & & & \\
Batch Size & 1024 & & & & & \\
\# of Epochs & 5000 & & & & & \\
$\alpha$ & 0.001 & & & & & \\
$\beta$ & 1000 & & & & & \\[0.5ex]
\end{tabular}
\end{center}
\end{table}
}

\def\tabletrDCVAECelebA
{
\begin{table}[h!]
\caption{U-Net trVAE detailed architecture used for CelebA dataset.}
\label{tab:trDCVAE_CelebA}
\begin{center}
\begin{tabular}{lllllll}
\multicolumn{1}{c}{\bf Name}  &\multicolumn{1}{c}{\bf Operation}&\multicolumn{1}{c}{\bf NoF/Kernel Dim.}&\multicolumn{1}{c}{\bf Dropout}&\multicolumn{1}{c}{\bf BN}&\multicolumn{1}{c}{\bf Activation}&\multicolumn{1}{c}{\bf Input}\\ \hline \\ 
input & - & (64, 64, 3) & $\times$ & $\times$ & - & - \\ 
conditions & - & 2 & $\times$ & $\times$ & - & - \\ 
FC-1 & FC & 128 & $\times$ & $\times$ & ReLU & conditions \\
FC-2 & FC & 1024 & 0.2 & $\surd$ & ReLU & FC-1  \\
FC-2\_reshaped & Reshape & (64, 64, 1) & $\times$ & $\times$ & $\times$ & FC-2 \\
Conv\_1 & Conv2D & (3, 3, 64, 4) & $\times$ & $\times$ & ReLU & [FC-2\_reshaped, input] \\
Conv\_2 & Conv2D & (3, 3, 64, 64) & $\times$ & $\times$ & ReLU & Conv\_1 \\
Pool\_1 & Pooling2D & $\times$ & $\times$ & $\times$ & $\times$ & Conv\_2 \\
Conv\_3 & Conv2D & (3, 3, 128, 64) & $\times$ & $\times$ & ReLU & Pool\_1 \\
Conv\_4 & Conv2D & (3, 3, 128, 128) & $\times$ & $\times$ & ReLU & Conv\_3 \\
Pool\_2 & Pooling2D & $\times$ & $\times$ & $\times$ & $\times$ & Conv\_4 \\
Conv\_5 & Conv2D & (3, 3, 256, 128) & $\times$ & $\times$ & ReLU & Pool\_2 \\
Conv\_6 & Conv2D & (3, 3, 256, 256) & $\times$ & $\times$ & ReLU & Conv\_5 \\
Conv\_7 & Conv2D & (3, 3, 256, 256) & $\times$ & $\times$ & ReLU & Conv\_6 \\
Pool\_3 & Pooling2D & $\times$ & $\times$ & $\times$ & $\times$ & Conv\_7 \\
Conv\_8 & Conv2D & (3, 3, 512, 256) & $\times$ & $\times$ & ReLU & Pool\_3 \\
Conv\_9 & Conv2D & (3, 3, 512, 512) & $\times$ & $\times$ & ReLU & Conv\_8 \\
Conv\_10 & Conv2D & (3, 3, 512, 512) & $\times$ & $\times$ & ReLU & Conv\_9 \\
Pool\_4 & Pooling2D & $\times$ & $\times$ & $\times$ & $\times$ & Conv\_10 \\
Conv\_11 & Conv2D & (3, 3, 512, 256) & $\times$ & $\times$ & ReLU & Pool\_4 \\
Conv\_12 & Conv2D & (3, 3, 512, 512) & $\times$ & $\times$ & ReLU & Conv\_11 \\
Conv\_13 & Conv2D & (3, 3, 512, 512) & $\times$ & $\times$ & ReLU & Conv\_12 \\
Pool\_4 & Pooling2D & $\times$ & $\times$ & $\times$ & $\times$ & Conv\_13 \\
flat & Flatten & $\times$ & $\times$ & $\times$ & $\times$ & Pool\_4 \\
FC-3 & FC & 1024 & $\times$ & $\times$ & ReLU & flat  \\
FC-4 & FC & 256 & 0.2 & $\times$ & ReLU & FC-3  \\
mean & FC & 60 & $\times$ & $\times$ & Linear & FC-4 \\
var & FC & 60 & $\times$ & $\times$ & Linear & FC-4 \\
z-sample & FC & 60 & $\times$ & $\times$ & Linear & [mean, var] \\
FC-5 & FC & 128 & $\times$ & $\times$ & ReLU & conditions \\
MMD & FC & 256 & $\times$ & $\surd$ & ReLU & [z-sample, FC-5] \\
FC-6 & FC & 1024 & $\times$ & $\times$ & ReLU & MMD \\
FC-7 & FC & 4096 & $\times$ & $\times$ & ReLU & FC-6 \\
FC-7\_reshaped & Reshape & $\times$ & $\times$ & $\times$ & FC-7 \\

Conv\_transp\_1 & Conv2D Transpose & (3, 3, 512, 512) & $\times$ & $\times$ & ReLU & FC-7\_reshaped \\
Conv\_transp\_2 & Conv2D Transpose & (3, 3, 512, 512) & $\times$ & $\times$ & ReLU & Conv\_transp\_1 \\
Conv\_transp\_3 & Conv2D Transpose & (3, 3, 512, 512) & $\times$ & $\times$ & ReLU & Conv\_transp\_2 \\
up\_sample\_1 & UpSampling2D & $\times$ & $\times$ & $\times$ & $\times$ & Conv\_transp\_3 \\

Conv\_transp\_4 & Conv2D Transpose & (3, 3, 512, 512) & $\times$ & $\times$ & ReLU & up\_sample\_1 \\
Conv\_transp\_5 & Conv2D Transpose & (3, 3, 512, 512) & $\times$ & $\times$ & ReLU & Conv\_transp\_4 \\
Conv\_transp\_6 & Conv2D Transpose & (3, 3, 512, 512) & $\times$ & $\times$ & ReLU & Conv\_transp\_5 \\
up\_sample\_2 & UpSampling2D & $\times$ & $\times$ & $\times$ & $\times$ & Conv\_transp\_6 \\

Conv\_transp\_7 & Conv2D Transpose & (3, 3, 128, 256) & $\times$ & $\times$ & ReLU & up\_sample\_2 \\
Conv\_transp\_8 & Conv2D Transpose & (3, 3, 128, 128) & $\times$ & $\times$ & ReLU & Conv\_transp\_7 \\
up\_sample\_3 & UpSampling2D & $\times$ & $\times$ & $\times$ & $\times$ & Conv\_transp\_8 \\

Conv\_transp\_9 & Conv2D Transpose & (3, 3, 64, 128) & $\times$ & $\times$ & ReLU & up\_sample\_3 \\
Conv\_transp\_10 & Conv2D Transpose & (3, 3, 64, 64) & $\times$ & $\times$ & ReLU & Conv\_transp\_9 \\
output & Conv2D Transpose & (1, 1, 3, 64) & $\times$ & $\times$ & ReLU & Conv\_transp\_10 \\
\hline
Optimizer & Adam & & & & &  \\
Learning Rate & 0.001 & & & & & \\
Leaky ReLU slope & 0.2 & & & & & \\
Batch Size & 1024 & & & & & \\
\# of Epochs & 5000 & & & & & \\
$\alpha$ & 0.001 & & & & & \\
$\beta$ & 1000 & & & & & \\[0.5ex]
\end{tabular}
\end{center}
\end{table}
}

\def\tablescGen{
\begin{table}[h!]
\caption{scGen detailed architecture.}
\label{tab:scGen}
\begin{center}
\begin{tabular}{lllllll}
\multicolumn{1}{c}{\bf Name}  &\multicolumn{1}{c}{\bf Operation}&\multicolumn{1}{c}{\bf NoF/Kernel Dim.}&\multicolumn{1}{c}{\bf Dropout}&\multicolumn{1}{c}{\bf BN}&\multicolumn{1}{c}{\bf Activation}&\multicolumn{1}{c}{\bf Input}\\ \hline \\ 
input & - & input\_dim & $\times$ & $\times$ & - & - \\
FC-1 & FC & 800 & 0.2 & $\surd$ & Leaky ReLU & input \\
FC-2 & FC & 800 & 0.2 & $\surd$ & Leaky ReLU & FC-1  \\
FC-3 & FC & 128 & 0.2 & $\surd$ & Leaky ReLU & FC-2  \\
mean & FC & 100 & $\times$ & $\times$ & Linear & FC-3 \\
var & FC & 100 & $\times$ & $\times$ & Linear & FC-3 \\
z & FC & 100 & $\times$ & $\times$ & Linear & [mean, var] \\
MMD & FC & 128 & 0.2 & $\surd$ & Leaky ReLU & z \\
FC-4 & FC & 800 & 0.2 & $\surd$ & Leaky ReLU & MMD \\
FC-5 & FC & 800 & 0.2 & $\surd$ & Leaky ReLU & FC-3 \\
output & FC & input\_dim & $\times$ & $\times$ & ReLU & FC-4 \\ 

\hline
Optimizer & Adam & & & & &  \\
Learning Rate & 0.001 & & & & & \\
Leaky ReLU slope & 0.2 & & & & & \\
Batch Size & 32 & & & & & \\
\# of Epochs & 300 & & & & & \\
$\alpha$ & 0.00050 & & & & & \\
$\beta$ & 100 & & & & & \\
$\eta$ & 100 & & & & & \\
\end{tabular}
\end{center}
\end{table}
}	

\def\tablescVI{
\begin{table}[h!]
\caption{scVI detailed architecture.}
\label{tab:scVI}
\begin{center}
\begin{tabular}{lllllll}
\multicolumn{1}{c}{\bf Name}  &\multicolumn{1}{c}{\bf Operation}&\multicolumn{1}{c}{\bf NoF/Kernel Dim.}&\multicolumn{1}{c}{\bf Dropout}&\multicolumn{1}{c}{\bf BN}&\multicolumn{1}{c}{\bf Activation}&\multicolumn{1}{c}{\bf Input}\\ \hline \\ 
input & - & input\_dim & $\times$ & $\times$ & - & - \\
conditions & - & 1 & $\times$ & $\times$ & - & - \\ 
FC-1 & FC & 128 & 0.2 & $\surd$ & ReLU & input \\
mean & FC & 10 & $\times$ & $\times$ & Linear & FC-1 \\
var & FC & 10 & $\times$ & $\times$ & Linear & FC-1 \\
z & FC & 10 & $\times$ & $\times$ & Linear & [mean, var] \\
FC-2 & FC & 128 & 0.2 & $\surd$ & ReLU & [z, conditions] \\
output & FC & input\_dim & $\times$ & $\times$ & ReLU & FC-2 \\ 
\hline
Optimizer & Adam & & & & &  \\
Learning Rate & 0.001 & & & & & \\
Batch Size & 128 & & & & & \\
\# of Epochs & 1000 & & & & & \\
$\alpha$ & 0.001 & & & & & \\
\end{tabular}
\end{center}
\end{table}
}

\def\tableMMDCVAE{
\begin{table}[h!]
\caption{MMD-CVAE detailed architecture.}
\label{tab:MMDCVAE}
\begin{center}
\begin{tabular}{lllllll}
\multicolumn{1}{c}{\bf Name}  &\multicolumn{1}{c}{\bf Operation}&\multicolumn{1}{c}{\bf NoF/Kernel Dim.}&\multicolumn{1}{c}{\bf Dropout}&\multicolumn{1}{c}{\bf BN}&\multicolumn{1}{c}{\bf Activation}&\multicolumn{1}{c}{\bf Input}\\ \hline \\ 
input & - & input\_dim & $\times$ & $\times$ & - & - \\
conditions & - & 1 & $\times$ & $\times$ & - & - \\ 
FC-1 & FC & 800 & 0.2 & $\surd$ & Leaky ReLU & [input, conditions] \\
FC-2 & FC & 800 & 0.2 & $\surd$ & Leaky ReLU & FC-1  \\
FC-3 & FC & 128 & 0.2 & $\surd$ & Leaky ReLU & FC-2  \\
mean & FC & 50 & $\times$ & $\times$ & Linear & FC-3 \\
var & FC & 50 & $\times$ & $\times$ & Linear & FC-3 \\
z-sample & FC & 50 & $\times$ & $\times$ & Linear & [mean, var] \\
MMD & FC & 128 & 0.2 & $\surd$ & Leaky ReLU & [z-sample, conditions] \\
FC-4 & FC & 800 & 0.2 & $\surd$ & Leaky ReLU & MMD \\
FC-5 & FC & 800 & 0.2 & $\surd$ & Leaky ReLU & FC-3 \\
output & FC & input\_dim & $\times$ & $\times$ & ReLU & FC-4 \\ 
\hline
Optimizer & Adam & & & & &  \\
Learning Rate & 0.001 & & & & & \\
Leaky ReLU slope & 0.2 & & & & & \\
Batch Size & 512 & & & & & \\
\# of Epochs & 500 & & & & & \\
$\alpha$ & 0.001 & & & & & \\
$\beta$ & 1 & & & & & \\
\end{tabular}
\end{center}
\end{table}
}	

\def\tableCVAE{
\begin{table}[h!]
\caption{CVAE detailed architecture.}
\label{tab:CVAE}
\begin{center}
\begin{tabular}{lllllll}
\multicolumn{1}{c}{\bf Name}  &\multicolumn{1}{c}{\bf Operation}&\multicolumn{1}{c}{\bf NoF/Kernel Dim.}&\multicolumn{1}{c}{\bf Dropout}&\multicolumn{1}{c}{\bf BN}&\multicolumn{1}{c}{\bf Activation}&\multicolumn{1}{c}{\bf Input}\\ \hline \\ 
input & - & input\_dim & $\times$ & $\times$ & - & - \\
conditions & - & 1 & $\times$ & $\times$ & - & - \\ 
FC-1 & FC & 800 & 0.2 & $\surd$ & Leaky ReLU & [input, conditions] \\
FC-2 & FC & 800 & 0.2 & $\surd$ & Leaky ReLU & FC-1  \\
FC-3 & FC & 128 & 0.2 & $\surd$ & Leaky ReLU & FC-2  \\
mean & FC & 50 & $\times$ & $\times$ & Linear & FC-3 \\
var & FC & 50 & $\times$ & $\times$ & Linear & FC-3 \\
z-sample & FC & 50 & $\times$ & $\times$ & Linear & [mean, var] \\
MMD & FC & 128 & 0.2 & $\surd$ & Leaky ReLU & [z-sample, conditions] \\
FC-4 & FC & 800 & 0.2 & $\surd$ & Leaky ReLU & MMD \\
FC-5 & FC & 800 & 0.2 & $\surd$ & Leaky ReLU & FC-3 \\
output & FC & input\_dim & $\times$ & $\times$ & ReLU & FC-4 \\ 
\hline
Optimizer & Adam & & & & &  \\
Learning Rate & 0.001 & & & & & \\
Leaky ReLU slope & 0.2 & & & & & \\
Batch Size & 512 & & & & & \\
\# of Epochs & 300 & & & & & \\
$\alpha$ & 0.001 & & & & & \\
\end{tabular}
\end{center}
\end{table}
}

\def\tableCycleGAN
{
\begin{table}[h!]
\caption{Style transfer GAN detailed architecture.}
\label{tab:CycleGAN}
\begin{center}
\begin{tabular}{lllllll}
\multicolumn{1}{c}{\bf Name}  &\multicolumn{1}{c}{\bf Operation}&\multicolumn{1}{c}{\bf NoF/Kernel Dim.}&\multicolumn{1}{c}{\bf Dropout}&\multicolumn{1}{c}{\bf BN}&\multicolumn{1}{c}{\bf Activation}&\multicolumn{1}{c}{\bf Input}\\ \hline \\ 
input & - & input\_dim & $\times$ & $\times$ & - & - \\ 
FC-1 & FC & 700 & 0.5 & $\surd$ & Leaky ReLU & input \\
FC-2 & FC & 100 & 0.5 & $\surd$ & Leaky ReLU & FC-1  \\
FC-3 & FC & 50 & 0.5 & $\surd$ & Leaky ReLU & FC-2 \\
FC-4 & FC & 100 & 0.5 & $\surd$ & Leaky ReLU & FC-3 \\
FC-5 & FC & 700 & 0.5 & $\surd$ & Leaky ReLU & FC-4 \\
generator\_out & FC & 6,998 & $\times$ & $\surd$ & ReLU & FC-5 \\ 
FC-6 & FC & 700 & 0.5 & $\surd$ & Leaky ReLU & generator\_out \\
FC-7 & FC & 100 & 0.5 & $\surd$ & Leaky ReLU & FC-6 \\
discriminator\_out & FC & 1 &  $\times$ & $\times$ & Sigmoid & FC-7 \\
Generator Optimizer & Adam & & & & &  \\
Discriminator Optimizer & Adam & & & & &  \\
\hline
Optimizer & Adam & & & & & \\
Learning Rate & 0.001 & & & & & \\
Leaky ReLU slope & 0.2 & & & & & \\
\# of Epochs & 1000 & & & & & \\ [0.5ex]
\end{tabular}
\end{center}
\end{table}
}

\def\tableSAUCIE{
\begin{table}[h!]
\caption{SAUCIE detailed architecture.}
\label{tab:SAUCIE}
\begin{center}
\begin{tabular}{lllllll}
\multicolumn{1}{c}{\bf Name}  &\multicolumn{1}{c}{\bf Operation}&\multicolumn{1}{c}{\bf NoF/Kernel Dim.}&\multicolumn{1}{c}{\bf Dropout}&\multicolumn{1}{c}{\bf BN}&\multicolumn{1}{c}{\bf Activation}&\multicolumn{1}{c}{\bf Input}\\ \hline \\ 
input & - & input\_dim & $\times$ & $\times$ & - & - \\
conditions & - & 1 & $\times$ & $\times$ & - & - \\ 
FC-1 & FC & 512 & $\times$ & $\surd$ & Leaky ReLU & [input, conditions] \\
FC-2 & FC & 256 & $\times$ & $\times$ & Leaky ReLU & FC-1  \\
FC-3 & FC & 128 & $\times$ & $\times$ & Leaky ReLU & FC-2  \\
FC-4 & FC & 20 & $\times$ & $\times$ & Leaky ReLU & FC-3 \\
FC-5 & FC & 128 & $\times$ & $\times$ & Leaky ReLU & FC-4 \\
FC-6 & FC & 256 & $\times$ & $\times$ & Leaky ReLU & FC-5 \\
FC-7 & FC & 512 & $\times$ & $\times$ & Leaky ReLU & FC-6 \\
output & FC & input\_dim & $\times$ & $\times$ & ReLU & FC-4 \\ 
\hline
Optimizer & Adam & & & & &  \\
Learning Rate & 0.001 & & & & & \\
Leaky ReLU slope & 0.2 & & & & & \\
Batch Size & 256 & & & & & \\
\# of Epochs & 1000 & & & & & \\
\end{tabular}
\end{center}
\end{table}
}	

\def\header{
\thispagestyle{empty}
\vspace*{-4.5em}\hfill \today\\[2em]
\noindent
\rule{\textwidth}{.1em}\\[.8em] 
{\rmfamily\bfseries\huge
Conditional out-of-sample generation for un-\\[.2em]paired data using trVAE
}\\[-.1em]
\rule{\textwidth}{.1em}\\[.5em]
M. Lotfollahi\textsuperscript{1,2},
Mohsen Naghipourfar \textsuperscript{1,4},
Fabian J.~Theis\textsuperscript{1,2,3$\dagger$}
\& F.~Alexander Wolf\textsuperscript{1$\ddagger$}
\\
\textbf{1}~Institute of Computational Biology, Helmholtz Center Munich, Neuherberg, Germany.
\\
\textbf{2}~School of Life Sciences Weihenstephan, Technical University of Munich, Munich, Germany
\\
\textbf{3}~Department of Mathematics, Technische Universität München, Munich, Germany.
\\
\textbf{4}~Department of Computer Engineering, Sharif University of Technology, Tehran, Iran.
\\
$\dagger$ fabian.theis@helmholtz-muenchen.de
$\ddagger$ alex.wolf@helmholtz-muenchen.de

\setlength{\parindent}{0pt}
\setlength{\parskip}{.5em}
\vspace{2em}
}

\begin{document}
\header

\begin{abstract}
While generative models have shown great success in generating high-dimensional samples conditional on low-dimensional descriptors (learning e.g. stroke thickness in MNIST, hair color in CelebA, or speaker identity in Wavenet), their generation out-of-sample poses fundamental problems. The conditional variational autoencoder (CVAE) as a simple conditional generative model does not explicitly relate conditions during training and, hence, has no incentive of learning a compact joint distribution across conditions. We overcome this limitation by matching their distributions using maximum mean discrepancy (MMD) in the decoder layer that follows the bottleneck. This introduces a strong regularization both for reconstructing samples within the same condition and for transforming samples across conditions, resulting in much improved generalization. We refer to the architecture as \emph{transformer} VAE (trVAE). Benchmarking trVAE on high-dimensional image and tabular data, we demonstrate higher robustness and higher accuracy than existing approaches. In particular, we show qualitatively improved predictions for cellular perturbation response to treatment and disease based on high-dimensional single-cell gene expression data, by tackling previously problematic minority classes and multiple conditions. For generic tasks, we improve Pearson correlations of high-dimensional estimated means and variances with their ground truths from 0.89 to 0.97 and 0.75 to 0.87, respectively.
\end{abstract}

\section{Introduction}
\label{sec:intro}

The task of generating high-dimensional samples $x$ conditional on a latent random vector $z$ and a categorical variable $s$ has established solutions \citep{mirza2014conditional,ren2016conditional}. The situation becomes more complicated if the support of $z$ is divided into different domains $d$ with different semantic meanings: say $d\in\{\text{men}, \text{women}\}$ and one is interested in out-of-sample generation of samples $x$ in a domain and condition $(d, s)$ that is not part of the training data. If one predicts how a given black-haired man would look with blonde hair, which we refer to as \emph{transforming} $x_{\text{men, black-hair}} \mapsto x_{\text{men, blonde-hair}}$, this becomes an out-of-sample problem if the training data does not have instances of blonde-haired men, but merely of blonde- and black-haired woman and black-haired men. In an application with higher relevance, there is strong interest in how untreated ($s=0$) humans ($d=0$) respond to drug treatment ($s=1$) based on training data from in vitro ($d=1$) and mice ($d=2$) experiments. Hence, the target domain of interest ($d=0$) does not offer training data for $s=1$, but only for $s=0$.

In the present paper, we suggest to address the challenge of transforming out-of-sample by regularizing the joint distribution across the categorical variable $s$ using maximum mean discrepancy (MMD) in the framework of a conditional variational autoencoder (CVAE) \citep{sohn15}. This produces a more compact representation of a distribution that displays high variance in the vanilla CVAE, which incentivizes learning of features across $s$ and results in more accurate out-of-sample prediction. MMD has proven successful in a variety of tasks. In particular, matching distributions with MMD in variational autoencoders \citep{kingma2013auto} has been put forward for unsupervised domain adaptation \citep{louizos16} or for learning statistically independent latent dimensions \citep{lopez18}. In supervised domain adaptation approaches, MMD-based regularization has been shown to be a viable strategy of learning label-predictive features with domain-specific information removed \citep{long15, tzeng14}.

The general idea of matching distributions across perturbed and control populations has been previously studied in the context of causal inference \citep{johansson16}, albeit not in the context of out-of-sample transformation and the CVAE. \citet{johansson16} showed how to improve counterfactual inference by learning representations that enforce similarity between perturbed and control using a linear discrepancy measure, mentioning MMD as an alternative metric.

Finally, in further related work, the out-of-sample transformation problem was addressed via hard-coded latent space vector arithmetics \citep{lotfollahi2019scgen} and histogram matching \citep{amodio2018out}. The approach of the present paper, however, introduces a data-driven end-to-end approach, which does not involve hard-coded elements and generalizes to more than one condition.

\section{Background}
\label{background}

\subsection{Variational autoencoder}

The motivation of the variational autoencoder (VAE) \citep{kingma2013auto} is to provide a neural-network based parametrization for maximizing the likelihood
\begin{equation}
	p_\theta(X \mid S) = \int p_\theta(X \mid Z, S) p_\theta(Z \mid S)dZ,
\end{equation}
where $X$ denotes a high-dimensional random variable, $S$ a random variable representing conditions, $\theta$ the model parameters, and $p_\theta(X \mid Z, S)$ the generative distribution that decodes $Z$ into $X$. Here and in the following we adapt the notation of \citet{lopez18} and summarize \citet{doersch16}.

To make the integral tractable by sampling values of $Z$ that are likely to produce values of $X$, one introduces an encoding distribution $q_\phi$, which can be related via 
\begin{multline}
\log p_\theta(X \mid S) - \KL{(q_\phi(Z|X, S)}||{p_\theta(Z|X, S))} \nonumber\\
= \mathbb{E}_{q_\phi(Z \mid X, S)}[\log p_\theta(X \mid Z, S)] - \KL{(q_\phi(Z|X, S)} || {p_\theta(Z|S))}.
\end{multline}
The right hand side of this equation provides the cost function for optimizing neural-network based parametrizations of $p_\theta$ and $q_\phi$. The left-hand side describes the likelihood subtracted by an error term.

The case in which $S \neq \emptyset$ is referred to as the conditional variational autoencoder (CVAE) \citep{sohn15}, and a straight-forward extension of the original framework \citep{kingma2013auto}, which treated $S \equiv \emptyset$. 

\subsection{Maximum-mean discrepancy}

Let $(\Omega, \mathcal{F}, \mathbb{P})$ be a probability space, $\mathcal{X}$ a separable metric space, $x: \Omega \rightarrow \mathcal{X}$ a random variable and $k: \mathcal{X} \times \mathcal{X} \rightarrow \mathbb{R}$ a continuous, bounded, positive semi-definite kernel with a corresponding reproducing kernel Hilbert space (RKHS) $\mathcal{H}$. Consider the kernel-based estimate of a distance between two distributions $p$ and $q$ over the random variables $X$ and $X'$. Such a distance, defined via the canonical distance between their $\mathcal{H}$-embeddings, is called the maximum mean discrepancy~\citep{gretton12} and denoted $l_\mathrm{MMD}(p, q)$, with an explicit expression:
\begin{align}
	\ell_{\mathrm{MMD}}(\*X, \*X') &= \frac{1}{n_0^2} \sum_{n, m} k(\*x_n, \*x_{m}) + \frac{1}{n_1^2} \sum_{n, m} k(\*x'_n, \*x'_m) - \frac{2}{n_0 n_1} \sum_{n, m} k(\*x_n, \*x'_m)\label{eq:mmd},
\end{align}
where the sums run over the number of samples $n_0$ and $n_1$ for $x$ and $x'$, respectively. Asymptotically, for a universal kernel such as the Gaussian kernel $k(x,x')=e^{-\gamma \| \*x - \*x' \|^2}$, $\ell_{\mathrm{MMD}}(\*X, \*X')$ is $0$ if and only if $p \equiv q$. For the implementation, we use multi-scale RBF kernels defined as:
\begin{equation}
    k(\*x,x')= \sum_{i=1}^{l}k(x,x',\gamma_{i})
\end{equation}
where $k(x,x',\gamma_{i})=e^{-\gamma_{i} \| \*x - \*x' \|^2}$ and $\gamma_{i}$ is a hyper-parameter.

Addressing the domain adaptation problem, the ``Variational Fair Autoencoder'' (VFAE) \citep{louizos16} uses MMD to match latent distributions $q_{\phi}(\*Z|s=0)$ and $q_{\phi}(\*Z| s=1)$ --- where $s$ denotes a domain --- by adapting the standard VAE cost function $\mathcal{L}_{\text{VAE}}$ according to
\begin{align}
    \mathcal{L}_{\text{VFAE}}(\phi, \theta; X, X',S,S') & = \mathcal{L}_{\text{VAE}}(\phi, \theta; X,S) + \mathcal{L}_{\text{VAE}}(\phi, \theta; X',S') -\beta \ell_{\mathrm{MMD}}({Z}_{s=0}, {Z'}_{s'=1}),
\end{align}
where $X$ and $X'$ are two high-dimensional observations with their respective conditions $S$ and $S'$.

In contrast to GANs \citep{goodfellow2014generative} whose training procedure is notoriously hard due to the minmax optimization problem, training models using MMD or Wasserstein distance metrics is comparatively simple \citep{li2015generative,pmlr-v70-arjovsky17a,Dziugaite:2015:TGN:3020847.3020875} as only a direct minimization of a simple loss is involved. It has been shown that MMD-based GANs have some advantages over Wasserstein GANs resulting in a simpler and faster-training algorithm with matching performance \citep{binkowski2018demystifying}. This motivated us to choose MMD as a metric for  regularizing distribution matching.

\trVAE

\section{Defining the transformer VAE}

Let us adapt the following notation for the transformation within a standard CVAE. High-dimensional observations $x$ and a scalar or low-dimensional condition $s$ are transformed using $f$ (encoder, corresponding to distribution $q_\phi$) and $g$ (decoder, corresponding to distribution $p_\theta$), which are parametrized by weight-sharing neural networks, and give rise to predictors $\hat{z}, \hat{y}$ and $\hat{x}$:
\begin{subequations}
\begin{align}
\hat{z} & = f(x, s) \\
\hat{y} & = g_1(\hat{z}, s) \label{g1}\\
\hat{x} & = g_{2}(\hat{y})
\end{align}
\end{subequations}
where we distinguished the first ($g_1$) and the remaining layers ($g_{2}$) of the decoder $g = g_{2} \circ g_{1}$ (Fig. \ref{fig:trVAE}).

While $z$ formally depends on $s$, it is commonly  empirically observed $Z \indep S$, that is, the representation $z$ is disentangled from the condition information $s$. By contrast, the original representation typically strongly covaries with $S$: $X \notindep S$. The observation can be explained by admitting that an efficient $z$-representation, suitable for minimizing reconstruction and regularization losses, should be as free as possible from information about $s$. Information about $s$ is directly and explicitly available to the decoder (\eqref{g1}), and hence, there is an incentive to optimize the parameters of $f$ to \emph{only} explain the variation in $x$ that is \emph{not} explained by $s$. Experiments below demonstrate that indeed, MMD regularization on the \emph{bottleneck layer} $z$ does not improve performance.

However, even if $z$ is completely free of variation from $s$, the $y$ representation has a strong $s$ component, $Y \notindep S$, which leads to a separation of $y_{s=1}$ and $y_{s=0}$ into different regions of their support $\mathcal{Y}$. In the standard CVAE, without any regularization of this $y$ representation, a highly varying, non-compact distribution emerges across different values of $s$ (Fig. \ref{fig:latent}). To compactify the distribution so that it displays only subtle, controlled differences, we impose MMD (\eqref{eq:mmd}) in the first layer of the decoder (Fig. \ref{fig:trVAE}). We assume that modeling $y$ in the same region of the support of $\mathcal{Y}$ across $s$ forces learning common features across $s$ where possible. The more of these common features are learned, the more accurately the transformation task will performed, and the higher are chances of successful out-of-sample generation.  Using one of the benchmark datasets introduced, below, we qualitatively illustrate the effect (Fig. \ref{fig:latent}).

During training time, all samples are passed to the model with their corresponding condition labels $(x_s, s)$. At prediction time, we pass $(x_{s=0}, s=0)$ to the encoder $f$ to obtain the latent representation $\hat z_{s=0}$. In the decoder $g$, we pass $(\hat z_{s=0}, s=1)$ and through that, let the model transform data to $\hat x_{s=1}$.

The cost function of trVAE derives directly from the standard CVAE cost function, as introduced in the backgrounds section,
\begin{align}
    \mathcal{L}_{\text{CVAE}}(\phi, \theta; X, S, \alpha, \eta) & = \eta\mathbb{E}_{q_\theta(Z|X, S)}\log(p_\phi(X | Z, S)) - \alpha \KL({q_\theta(Z|X, S)}||p_\phi(Z|X, S)).
\label{eq:cvae}
\end{align}
Consistent with the above, let $\hat{y}_{s=0} = g_1(f(x, s=0), s=0)$ and $\hat{y}_{s=1} = g_1(f(x', s=1), s=1)$. Through duplicating the cost function for $X'$ and adding an MMD term, the loss of trVAE becomes:
\begin{align}
    \mathcal{L}_{\text{trVAE}}(\phi, \theta; X, X', S, S', \alpha, \eta,\beta)
    = &~ \mathcal{L}_{\text{CVAE}}(\phi, \theta; X, S, \alpha, \eta) \nonumber\\
    & + \mathcal{L}_{\text{CVAE}}(\phi, \theta; X', S', \alpha, \eta) \nonumber\\
    & -\beta \ell_{\mathrm{MMD}}(\hat{Y}_{s=0}, \hat{Y}_{s'=1}).
\end{align}

\latent

\section{Experiments}
\label{sec:Experiments}

We demonstrate the advantages of an MMD-regularized first layer of the decoder by benchmarking versus a variety of existing methods and alternatives:
\begin{itemize}
    \item Vanilla CVAE \citep{sohn15}
    \item CVAE with MMD on bottleneck (MMD-CVAE), similar to VFAE \citep{louizos16}
    \item MMD-regularized autoencoder \citep{dziugaite2015training,Amodio237065}
    \item CycleGAN \citep{zhu17}
    \item scGen, a VAE combined with vector arithmetics \citep{lotfollahi2019scgen}
    \item scVI, a CVAE with a  negative binomial output distribution \citep{lopez2018deep}
\end{itemize}

First, we demonstrate trVAE's basic out-of-sample style transfer capacity on two established image datasets, on a qualitative level. We then address quantitative comparisons of challenging benchmarks with clear ground truth, predicting the effects of biological perturbation based on high-dimensional structured data. We used convolutional layers for imaging examples in section \ref{sc:image} and fully connected layers for single-cell gene expression datasets in sections \ref{sc:haber} and \ref{sc:Kang}. The optimal hyper-parameters for each application were chosen by using  a parameter gird-search for each model. The detailed hyper-parameters for different models are reported in tables \ref{tab:trDCVAE_MNIST}-\ref{tab:SAUCIE} in appendix \ref{supp:tables}.

\subsection{MNIST and CelebA style transformation}
\label{sc:image}

Here, we use Morpho-MNIST \citep{castro2018morphomnist}, which contains 60,000 images each of "normal" and "transformed" digits, which are drawn with a thinner and thicker stroke. For training, we used all normal-stroke data. Hence, the training data covers all domains ($d\in\{0, 1, 2, \dots, 9\}$) in the normal stroke condition ($s=0$). In the transformed conditions (thin and thick strokes, $s\in\{1, 2\}$), we only kept domains $d\in\{1, 3, 6, 7\}$.

We train a convolutional trVAE in which we first encode the stroke width via two fully-connected layers with 128 and 784 features, respectively. Next, we reshape the 784-dimensional into 28*28*1 images and add them as another channel in the image. Such trained trVAE faithfully transforms digits of normal stroke to digits of thin and thicker stroke to the out-of-sample domains (Fig. \ref{fig:mnist})

\MNIST

Next, we apply trVAE to CelebA \citep{liu2015celeba}, which contains 202,599 images of celebrity faces with 40 binary attributes for each image. We focus on the task of learning a transformation that turns a non-smiling face into a smiling face. We kept the smiling ($s$) and gender ($d$) attributes and trained the model with images from both smiling and non-smiling men but only with non-smiling women.

In this case, we trained a deep convolutional trVAE with a U-Net-like architecture \citep{Ronneberger_2015}. We encoded the binary condition labels as in the Morpho-MNIST example and fed them as an additional channel in the input.

\CelebA

Predicting out-of-sample, trVAE successfully transforms non-smiling faces of women to smiling faces while preserving most aspects of the original image (Fig. \ref{fig:celeba}). In addition to showing the model's capacity to handle more complex data, this example demonstrates the flexibility of the the model adapting to well-known architectures like U-Net in the field.

\subsection{Infection response}
\label{sc:haber}
\Haber

Accurately modeling cell response to perturbations is a key question in computational biology. Recently, neural network models have been proposed for out-of-sample predictions of high-dimensional tabular data that quantifies gene expression of single-cells \citep{lotfollahi2019scgen,amodio2018out}. However, these models are not trained on the task relying instead on hard-coded transformations and cannot handle more than two conditions.

We evaluate trVAE on a single-cell gene expression dataset that characterizes the gut \citep{Haber} after Salmonella or Heligmosomoides polygyrus (H. poly) infections, respectively. For this, we closely follow the benchmark as introduced in \citep{lotfollahi2019scgen}. The dataset contains eight different cell types in four conditions: control or healthy cells (n=3,240), H.Poly infection a after three days (H.Poly.Day3, n=2,121), H.poly infection after 10 days (H.Poly.Day10, n=2,711) and salmonella infection (n=1,770) (Fig. \ref{fig:Haber}a). The normalized gene expression data has 1,000 dimensions corresponding to 1,000 genes. Since three of the benchmark models are only able to handle two conditions, we only included the control and H.Poly.Day10 conditions for model comparisons. In this setting, we hold out Tuft infected cells for training and validation, as these consitute the hardest case for out-of-sample generalization (least shared features, few training data).

Figure \ref{fig:Haber}b-c shows trVAE accurately predicts the mean and variance for high-dimensional gene expression in Tuft cells. We compared the distribution of \textit{Defa24}, the gene with the highest change after H.poly infection in Tuft cells, which shows trVAE provides better estimates for mean and variance compared to other models. Moreover, trVAE outperforms other models also when quantifying the correlation of the predicted 1,000 dimensional $x$ with its ground truth (Fig. \ref{fig:Haber}e). In particular, we note that the MMD regularization on the \emph{bottleneck layer} of the CVAE does not improve performance, as argued above.

In order to show our model is able to handle multiple conditions, we performed another experiment with all three conditions included. We trained trVAE holding out each of the eight cells types in all perturbed conditions. Figure \ref{fig:Haber}f shows trVAE can accurately predict all cell types in each perturbed condition, in contrast to existing models.

\subsection{Stimulation response}

\label{sc:Kang}

\Kang

Similar to modeling infection response as above, we benchmark on another single-cell gene expression dataset consisting of 7,217 IFN-$\beta$ stimulated and 6,359 control peripheral blood mononuclear cells (PBMCs) from eight different human Lupus patients \citep{kang2018multiplexed}. The stimulation with IFN-$\beta$ induces dramatic changes in the transcriptional profiles of immune cells, which causes big shifts between control and stimulated cells (Fig. \ref{fig:Kang}a). We studied the out-of-sample prediction of natural killer (NK) cells held out during the training of the model.

trVAE accurately predicts mean (Fig. \ref{fig:Kang}b) and variance (Fig. \ref{fig:Kang}c) for all genes in the held out NK cells. In particular, genes strongly responding to IFN-$\beta$ (highlighted in red in Fig. \ref{fig:Kang}b-c) are well captured. An effect of applying IFN-$\beta$ is an increase in ISG15 for NK cells, which the model never sees during training. trVAE predicts this change by increasing the expression of ISG15 as observed in real NK cells (Fig. \ref{fig:Kang}d). A cycle GAN and an MMD-regularized auto-encoder (SAUCIE) and other models yield less accurate results than our model. Comparing the correlation of predicted mean and variance of gene expression for all dimensions of the data, we find trVAE performs best (Fig. \ref{fig:Kang}e).

\section{Discussion}

By arguing that the vanilla CVAE yields representations in the first layer following the bottleneck that vary strongly across categorical conditions, we introduced an MMD regularization that forces these representations to be similar across conditions. The resulting model (trVAE) outperforms existing modeling approaches on benchmark and real-world data sets.

Within the bottleneck layer, CVAEs already display a well-controlled behavior, and regularization does not improve performance. Further regularization at later layers might be beneficial but is numerically costly and unstable as representations become high-dimensional. However, we have not yet systematically investigated this and leave it for future studies.

Further future work will concern the application of trVAE on larger and more data, focusing on interaction effects among conditions. For this, an important application is the study of drug interaction effects, as previously noted by \citet{amodio2018out}. Future conceptual investigations concern establishing connections to causal-inference-inspired models beyond \citep{johansson16} such as CEVAE \citep{louizos2017causal}, establishing further that faithful modeling of an interventional distribution can be re-framed as successful perturbation effect prediction across domains.

\section{Acknowledgements}

We are grateful to Anna Klimovskaia for pointing us to the highly-related reference of Johansson \citep{johansson16} after submission to arXiv. We thank Romain Lopez for making us aware of sentences in our background section that were copied from the background section of \citet{lopez18}. These sentences were copied in an early brainstorming phase of the present manuscript and accidentally remained when writing the paper four months later. FAW is responsible for this mistake, acted assuming the sentences were due to a coauthor and not due to \citet{lopez18}, and is profoundly sorry.

\bibliographystyle{iclr2020_conference}
\bibliography{iclr2020_conference}

\begin{thebibliography}{29}
\providecommand{\natexlab}[1]{#1}
\providecommand{\url}[1]{\texttt{#1}}
\expandafter\ifx\csname urlstyle\endcsname\relax
  \providecommand{\doi}[1]{doi: #1}\else
  \providecommand{\doi}{doi: \begingroup \urlstyle{rm}\Url}\fi

\bibitem[Amodio et~al.(2018)Amodio, van Dijk, Montgomery, Wolf, and
  Krishnaswamy]{amodio2018out}
Matthew Amodio, David van Dijk, Ruth Montgomery, Guy Wolf, and Smita
  Krishnaswamy.
\newblock Out-of-sample extrapolation with neuron editing.
\newblock \emph{arXiv:1805.12198}, 2018.

\bibitem[Amodio et~al.(2019)Amodio, van Dijk, Srinivasan, Chen, Mohsen, Moon,
  Campbell, Zhao, Wang, Venkataswamy, Desai, Ravi, Kumar, Montgomery, Wolf, and
  Krishnaswamy]{Amodio237065}
Matthew Amodio, David van Dijk, Krishnan Srinivasan, William~S Chen, Hussein
  Mohsen, Kevin~R. Moon, Allison Campbell, Yujiao Zhao, Xiaomei Wang,
  Manjunatha Venkataswamy, Anita Desai, V.~Ravi, Priti Kumar, Ruth Montgomery,
  Guy Wolf, and Smita Krishnaswamy.
\newblock Exploring single-cell data with deep multitasking neural networks.
\newblock \emph{bioRxiv}, 2019.
\newblock \doi{10.1101/237065}.

\bibitem[Arjovsky et~al.(2017)Arjovsky, Chintala, and
  Bottou]{pmlr-v70-arjovsky17a}
Martin Arjovsky, Soumith Chintala, and L{\'e}on Bottou.
\newblock {W}asserstein generative adversarial networks.
\newblock In Doina Precup and Yee~Whye Teh (eds.), \emph{Proceedings of the
  34th International Conference on Machine Learning}, volume~70 of
  \emph{Proceedings of Machine Learning Research}, pp.\  214--223,
  International Convention Centre, Sydney, Australia, 06--11 Aug 2017. PMLR.

\bibitem[Bi{\'n}kowski et~al.(2018)Bi{\'n}kowski, Sutherland, Arbel, and
  Gretton]{binkowski2018demystifying}
Miko{\l}aj Bi{\'n}kowski, Dougal~J Sutherland, Michael Arbel, and Arthur
  Gretton.
\newblock Demystifying mmd gans.
\newblock \emph{arXiv:1801.01401}, 2018.

\bibitem[Castro et~al.(2018)Castro, Tan, Kainz, Konukoglu, and
  Glocker]{castro2018morphomnist}
Daniel~C. Castro, Jeremy Tan, Bernhard Kainz, Ender Konukoglu, and Ben Glocker.
\newblock {Morpho-MNIST}: Quantitative assessment and diagnostics for
  representation learning.
\newblock 2018.

\bibitem[Doersch(2016)]{doersch16}
Carl Doersch.
\newblock Tutorial on variational autoencoders.
\newblock \emph{arXiv:1606.05908}, 2016.

\bibitem[Dziugaite et~al.(2015{\natexlab{a}})Dziugaite, Roy, and
  Ghahramani]{Dziugaite:2015:TGN:3020847.3020875}
Gintare~Karolina Dziugaite, Daniel~M. Roy, and Zoubin Ghahramani.
\newblock Training generative neural networks via maximum mean discrepancy
  optimization.
\newblock In \emph{Proceedings of the Thirty-First Conference on Uncertainty in
  Artificial Intelligence}, UAI'15, pp.\  258--267, Arlington, Virginia, United
  States, 2015{\natexlab{a}}. AUAI Press.

\bibitem[Dziugaite et~al.(2015{\natexlab{b}})Dziugaite, Roy, and
  Ghahramani]{dziugaite2015training}
Gintare~Karolina Dziugaite, Daniel~M Roy, and Zoubin Ghahramani.
\newblock Training generative neural networks via maximum mean discrepancy
  optimization.
\newblock \emph{arXiv:1505.03906}, 2015{\natexlab{b}}.

\bibitem[Goodfellow et~al.(2014)Goodfellow, Pouget-Abadie, Mirza, Xu,
  Warde-Farley, Ozair, Courville, and Bengio]{goodfellow2014generative}
Ian Goodfellow, Jean Pouget-Abadie, Mehdi Mirza, Bing Xu, David Warde-Farley,
  Sherjil Ozair, Aaron Courville, and Yoshua Bengio.
\newblock Generative adversarial nets.
\newblock In \emph{Advances in neural information processing systems}, pp.\
  2672--2680, 2014.

\bibitem[Gretton et~al.(2012)Gretton, Borgwardt, Rasch, Sch{\"o}lkopf, and
  Smola]{gretton12}
Arthur Gretton, Karsten~M Borgwardt, Malte~J Rasch, Bernhard Sch{\"o}lkopf, and
  Alexander Smola.
\newblock A kernel two-sample test.
\newblock \emph{Journal of Machine Learning Research}, 13:\penalty0 723--773,
  2012.

\bibitem[Haber et~al.(2017)Haber, Biton, Rogel, Herbst, Shekhar, Smillie,
  Burgin, Delorey, Howitt, Katz, Tirosh, Beyaz, Dionne, Zhang, Raychowdhury,
  Garrett, Rozenblatt-Rosen, Shi, Yilmaz, Xavier, and Regev]{Haber}
Adam~L. Haber, Moshe Biton, Noga Rogel, Rebecca~H. Herbst, Karthik Shekhar,
  Christopher Smillie, Grace Burgin, Toni~M. Delorey, Michael~R. Howitt, Yarden
  Katz, Itay Tirosh, Semir Beyaz, Danielle Dionne, Mei Zhang, Raktima
  Raychowdhury, Wendy~S. Garrett, Orit Rozenblatt-Rosen, Hai~Ning Shi, Omer
  Yilmaz, Ramnik~J. Xavier, and Aviv Regev.
\newblock A single-cell survey of the small intestinal epithelium.
\newblock \emph{Nature}, 551:\penalty0 333, 2017.

\bibitem[Johansson et~al.(2016)Johansson, Shalit, and Sontag]{johansson16}
Fredrik Johansson, Uri Shalit, and David Sontag.
\newblock Learning representations for counterfactual inference.
\newblock In \emph{International conference on machine learning}, pp.\
  3020--3029, 2016.

\bibitem[Kang et~al.(2018)Kang, Subramaniam, Targ, Nguyen, Maliskova, McCarthy,
  Wan, Wong, Byrnes, Lanata, et~al.]{kang2018multiplexed}
Hyun~Min Kang, Meena Subramaniam, Sasha Targ, Michelle Nguyen, Lenka Maliskova,
  Elizabeth McCarthy, Eunice Wan, Simon Wong, Lauren Byrnes, Cristina~M Lanata,
  et~al.
\newblock Multiplexed droplet single-cell rna-sequencing using natural genetic
  variation.
\newblock \emph{Nature biotechnology}, 36\penalty0 (1):\penalty0 89, 2018.

\bibitem[Kingma \& Welling(2013)Kingma and Welling]{kingma2013auto}
Diederik~P Kingma and Max Welling.
\newblock Auto-encoding variational bayes.
\newblock \emph{arXiv:1312.6114}, 2013.

\bibitem[Li et~al.(2015)Li, Swersky, and Zemel]{li2015generative}
Yujia Li, Kevin Swersky, and Rich Zemel.
\newblock Generative moment matching networks.
\newblock In \emph{International Conference on Machine Learning}, pp.\
  1718--1727, 2015.

\bibitem[Liu et~al.(2015)Liu, Luo, Wang, and Tang]{liu2015celeba}
Ziwei Liu, Ping Luo, Xiaogang Wang, and Xiaoou Tang.
\newblock Deep learning face attributes in the wild.
\newblock In \emph{Proceedings of International Conference on Computer Vision
  (ICCV)}, December 2015.

\bibitem[Long et~al.(2015)Long, Cao, Wang, and Jordan]{long15}
Mingsheng Long, Yue Cao, Jianmin Wang, and Michael~I Jordan.
\newblock Learning transferable features with deep adaptation networks.
\newblock \emph{arXiv:1502.02791}, 2015.

\bibitem[Lopez et~al.(2018{\natexlab{a}})Lopez, Regier, Cole, Jordan, and
  Yosef]{lopez2018deep}
Romain Lopez, Jeffrey Regier, Michael~B. Cole, Michael~I. Jordan, and Nir
  Yosef.
\newblock Deep generative modeling for single-cell transcriptomics.
\newblock \emph{Nature Methods}, 15\penalty0 (12):\penalty0 1053--1058,
  2018{\natexlab{a}}.

\bibitem[Lopez et~al.(2018{\natexlab{b}})Lopez, Regier, Jordan, and
  Yosef]{lopez18}
Romain Lopez, Jeffrey Regier, Michael~I Jordan, and Nir Yosef.
\newblock Information constraints on auto-encoding variational bayes.
\newblock In \emph{Advances in Neural Information Processing Systems}, pp.\
  6114--6125, 2018{\natexlab{b}}.

\bibitem[Lotfollahi et~al.(2019)Lotfollahi, Wolf, and
  Theis]{lotfollahi2019scgen}
Mohammad Lotfollahi, F~Alexander Wolf, and Fabian~J Theis.
\newblock {scGen} predicts single-cell perturbation responses.
\newblock \emph{Nature methods}, 16\penalty0 (8):\penalty0 715, 2019.

\bibitem[Louizos et~al.(2015)Louizos, {Swersky}, {Li}, {Welling}, and
  {Zemel}]{louizos16}
Christos Louizos, Kevin {Swersky}, Yujia {Li}, Max {Welling}, and Richard
  {Zemel}.
\newblock The variational fair autoencoder.
\newblock \emph{arXiv:1511.00830}, 2015.

\bibitem[Louizos et~al.(2017)Louizos, Shalit, Mooij, Sontag, Zemel, and
  Welling]{louizos2017causal}
Christos Louizos, Uri Shalit, Joris~M Mooij, David Sontag, Richard Zemel, and
  Max Welling.
\newblock Causal effect inference with deep latent-variable models.
\newblock In \emph{Advances in Neural Information Processing Systems}, pp.\
  6446--6456, 2017.

\bibitem[{McInnes} et~al.(2018){McInnes}, {Healy}, and
  {Melville}]{2018arXivUMAP}
L.~{McInnes}, J.~{Healy}, and J.~{Melville}.
\newblock {UMAP: Uniform Manifold Approximation and Projection for Dimension
  Reduction}.
\newblock \emph{arXiv:1802.03426}, 2018.

\bibitem[Mirza \& Osindero(2014)Mirza and Osindero]{mirza2014conditional}
Mehdi Mirza and Simon Osindero.
\newblock Conditional generative adversarial nets.
\newblock \emph{arXiv:1411.1784}, 2014.

\bibitem[Ren et~al.(2016)Ren, Zhu, Li, and Luo]{ren2016conditional}
Yong Ren, Jun Zhu, Jialian Li, and Yucen Luo.
\newblock Conditional generative moment-matching networks.
\newblock In \emph{Advances in Neural Information Processing Systems}, pp.\
  2928--2936, 2016.

\bibitem[Ronneberger et~al.(2015)Ronneberger, Fischer, and
  Brox]{Ronneberger_2015}
Olaf Ronneberger, Philipp Fischer, and Thomas Brox.
\newblock U-net: Convolutional networks for biomedical image segmentation.
\newblock \emph{Medical Image Computing and Computer-Assisted Intervention –
  MICCAI 2015}, pp.\  234–241, 2015.

\bibitem[Sohn et~al.(2015)Sohn, Lee, and Yan]{sohn15}
Kihyuk Sohn, Honglak Lee, and Xinchen Yan.
\newblock Learning structured output representation using deep conditional
  generative models.
\newblock In \emph{Advances in Neural Information Processing Systems 28}, pp.\
  3483--3491. 2015.

\bibitem[Tzeng et~al.(2014)Tzeng, Hoffman, Zhang, Saenko, and Darrell]{tzeng14}
Eric Tzeng, Judy Hoffman, Ning Zhang, Kate Saenko, and Trevor Darrell.
\newblock Deep domain confusion: Maximizing for domain invariance.
\newblock \emph{arXiv:1412.3474}, 2014.

\bibitem[Zhu et~al.(2017)Zhu, Park, Isola, and Efros]{zhu17}
Jun-Yan Zhu, Taesung Park, Phillip Isola, and Alexei~A Efros.
\newblock Unpaired image-to-image translation using cycle-consistent
  adversarial networks.
\newblock In \emph{IEEE International Conference on Computer Vision (ICCV)},
  2017.

\end{thebibliography}
\newpage
\appendix

\section{hyper-parameters}
\label{supp:tables}

\tabletrDCVAEMNIST
\tabletrDCVAECelebA
\tabletrVAE
\tablescGen
\tableCVAE
\tableMMDCVAE
\tableCycleGAN
\tablescVI
\tableSAUCIE
\end{document}